\documentclass[conference]{IEEEtran}

\usepackage[english]{babel}
\usepackage[utf8x]{inputenc}
\usepackage[T1]{fontenc}
\usepackage{authblk}

\usepackage{amsmath}
\usepackage{graphicx, subfigure}
\usepackage[colorinlistoftodos]{todonotes}
\usepackage[colorlinks=true, allcolors=blue]{hyperref}

\usepackage[numbers]{natbib}

\usepackage{multirow}
\usepackage{colortbl}

\title{Offline and Online Deep Learning for Image Recognition}

\author{Nguyen Huu Phong and Bernardete Ribeiro}
\affil{
 CISUC -- Department of Informatics Engineering\\
 University of Coimbra, Polo II, Pinhal de Marrocos, \\3030--290 Coimbra, Portugal\\
 \{phong,bribeiro\}@dei.uc.pt
}
\date{}
\begin{document}
\maketitle

\begin{abstract}
Image recognition using Deep Learning has been evolved for decades though advances in the field through different settings is still a challenge. In this paper, we present our findings in searching for better image classifiers in offline and online environments. We resort to Convolutional Neural Network and its variations of fully connected Multi-layer Perceptron. Though still preliminary, these results are encouraging and may provide a better understanding about the field and directions toward future works.
\end{abstract}

\providecommand{\keywords}[1]{\textbf{\textit{Keywords---}} #1}

\renewcommand{\thesection}{\Roman{section}} 
\renewcommand\figurename{Fig.}
  
\section{Introduction}
\label{sec:intro}
Recent years have seen a re-appearance of Deep Learning from academy to business area. In academy, the technique has achieved significant higher classification accuracy on competitions such as image recognition \cite{imagenet2012} and speech recognition \cite{microsoft2016}. These results inspired by the previous works of LeCun, Bengio and Hilton on Deep Learning were catapulted with the availability of GPU and BigData \cite{lecun2015}. In business, Google self-driving cars have been tested in large cities and accumulated hundred years of human driving experience \cite{google2017}. Uber also made a breakthrough in public service as the first company to offer self-driving cars \cite{nytimes2017}. Deep Learning is being seen in Natural Language Processing e.g. Apple Siri, Google Now and Amazon Alexa which offer voice recognition services to assist customers in searching information. We believe that one of the next steps is Natural Language Understanding (NLU) for which much achievement is expected soon.

Before the arrival of Deep Learning in image classification, the field has evolved through several stages from Linear Classifier to Support Vector Machine and Neural Networks. These methods commonly require selection of features that eventually needs involvement of experts in particular fields. Deep Learning  on the other hand can choose the best feature automatically \cite{lecun2015}.

In this article, we employ MNIST and Cifar 10 –  standards for digit recognition and image recognition as testbeds for our classifiers. Table \ref{tab:table1} shows a summary of performances of MNIST based on several classifiers. The earliest one made by LeCun (also a main contributor of MNIST) with one layer neural network produced $12\%$ in terms of error rate \cite{Lecun1998}. Since then, several researches have been done to improve the performance. For example, authors in \cite{ebrahimzadeh2014} applied a concept of Support Vector Machine (SVM) that reduces error rate to $2.75\%$. In addition, authors in \cite{keysers2007} decreased the error rate by $5$ times. With the recent popularization of the Convolutional Neural Network, researchers were able to archive lower error rates with deeper layers of convolution as seen in references \cite{li2016,ciresan2011,cirecsan2012,wan2013}. At the time of this writing, the best error is $0.21$. For Cifar 10, the best error rate is $3.74$~\cite{graham2014fractional}.

The rest of this paper is organized as follows.  In Section~\ref{sec:contribution} we highlight our main contribution. In Section~\ref{sec:offline}, we deal with offline implementation and setups for comparisons of the shallow and deep classifiers in MNIST dataset. We discuss the results in Section~\ref{sec:results}. We also setup other online deep learning experiments with Cifar 10 in Section~\ref{sec:online}. Finally, we summarise our findings and approaches toward future work in Section~\ref{sec:conclusion}.

\begin{table}
\centering
\caption{\label{tab:table1}Classifiers' Performance on MNIST.}
\begin{tabular}{l|c|r}
\# Ref & Classifier & Error Rate (\%) \\\hline
\cite{Lecun1998} & 1 Layer NN & 12 \\
\cite{ebrahimzadeh2014} & SVM & 2.75 \\
\cite{keysers2007} & KNN with IDM & 0.54 \\
\cite{li2016} & Deep CNN & $0.47\pm0.05$ \\
\cite{ciresan2011} & 7 CNN & $0.27\pm0.02$ \\
\cite{cirecsan2012} & 35 CNN & 0.23 \\
\cite{wan2013} & DropConnect NN & 0.21 \\
\end{tabular}
\end{table}

\section{Contribution}
\label{sec:contribution}
In this research, we present our findings for a better performance in image recognition in offline and online settings. We have setup a development framework for performing offline image recognition. We also looked for the best setting in Multi-layer Perceptron and compare with Convolution Neural Networks. Moreover, in online setting, we also tried to find the most efficient architecture. Even though still preliminary, these results are very promising and provide approaches toward future exploration.

\section{Offline Implementation}
\label{sec:offline}
In this section, we discuss the benchmark dataset, the development framework as well as the setups of Multi-layer Perceptron and Convolutional Neural Networks.

\subsection{MNIST}
\begin{figure}[htb!]
\centering
\includegraphics[width=0.5\textwidth]{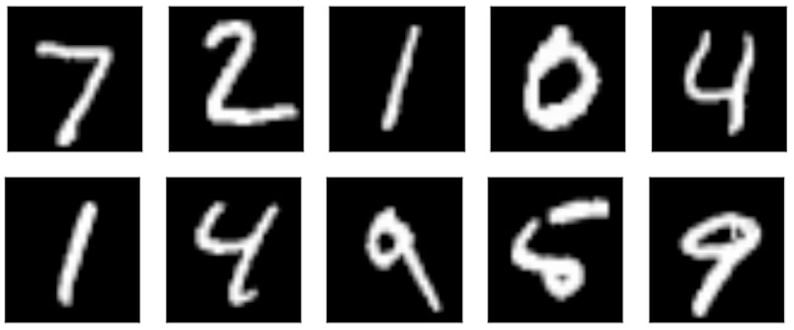}
\caption{\label{fig:visualization}Visualization of first ten digits MNIST.}
\end{figure}

We obtain dataset from MNIST which is a Modified version of United States' National Institute of Standards and Technology. The data has a training set of $60000$ samples and a testing set of $10000$ samples. The training and testing sets each include scanned handwritten images and desired outputs. These images were rescaled into $20\times 20$ pixel box and then centered in $28\times 28$ pixel field. Each digit in an image represents gray level ranging from 0 to 255 where 0 means white color and 255 means black color \cite{Lecun1998}. The visual figure of the first ten digits in the training set is shown in Fig. \ref{fig:visualization}.

\subsection{Development Framework}
\begin{figure}
\centering
\includegraphics[width=0.5\textwidth]{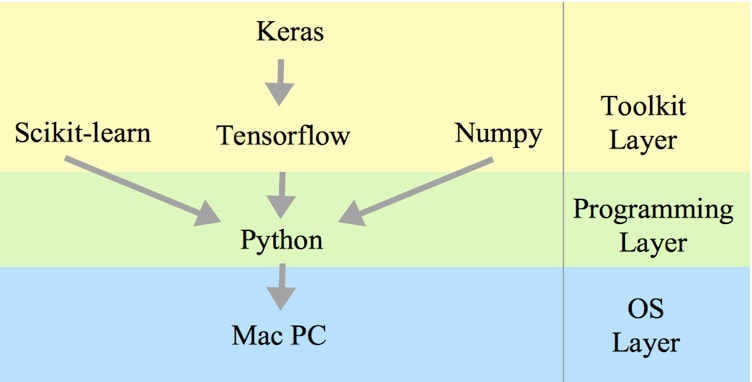}
\caption{\label{fig:development}Development Framework}
\end{figure}

In order to perform image recognition with deep learning, we setup a development framework which includes three layers, namely, OS Layer, Programming Layer, and Toolkit Layer as depicted in Fig. \ref{fig:development}. In the first layer, we perform our experiments on a Macbook Pro (Intel 2.7 GHz). In the programming layer, we choose Python since this language is one of the most popular programming languages in scientific community and the other reason is that Python operates very fast in runtime. In the toolkit layer, we build our image recognition surrounding Tensorflow library which is a deep learning library and has been supported by Google Inc since 2015. Besides of Tensorflow, there are also several deep learning libraries e.g. Theano, Torch and Deeplearning4j. For convenience of dealing with different libraries, we decided to use Keras on top of Tensorflow.

\subsection{Setup for MLP}
\begin{figure}[htb!]
\centering
\includegraphics[width=0.5\textwidth]{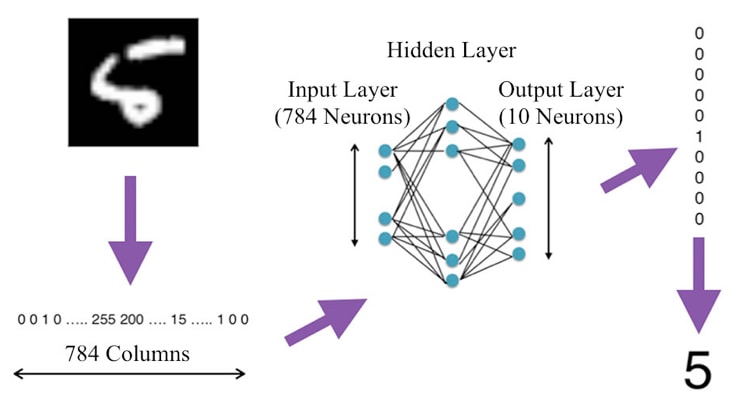}
\caption{\label{fig:processmlp}Process of classifying a sample digit 5 using an MLP}
\end{figure}

This section deals with how we perform training and testing of the datasets on MLP. Fig. \ref{fig:processmlp} shows the process of these steps. First of all, as mentioned in the previous section, each image encoded in an array of 28 rows and 28 columns is permuted into a vector of 784 columns. Then the vector will be used as input of our Neural Networks. After that, the data is processed in a fully connected MLP. In the output layer, we set 10 neurons for decoding 10 digits from 0 to 9. For example, the binary 0000000001 would represent digit 0. The binary 0000000010 would represent digit 1 so on and so forth.

The essential of an MLP is computing weights via Equation \ref{eq:1} where $w_i$ is the weight of the neuron $i$, $\eta$ is the learning rate, $t$,$o$ and $x$ are target, output and input respectively.
\begin{gather}
  w_i \leftarrow w_i+\eta(t-o)x\label{eq:1}
\end{gather}
\subsection{Setup for CNN}
\begin{figure}
\centering
\includegraphics[width=0.5\textwidth]{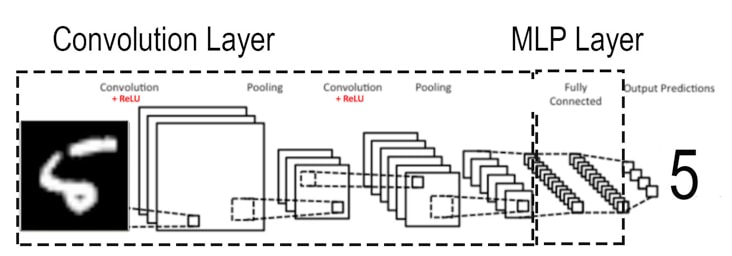}
\caption{\label{fig:processcnn}Process of classifying a sample digit 5 using an CNN (Adapted from \cite{lecun2015})}
\end{figure}

The structure of the Convolutional Neural Network is shown in Fig. \ref{fig:processcnn}. As we can see, this structure includes a convolutional layer in addition to multi-layer perception layers. The convolutional layer may include one or more combinations of convolution, pooling and ReLU stages. A unit employing the rectifier activation function is called a rectified linear unit (ReLU). While convolution performs as feature extraction, pooling and ReLu reduce the dimension of the convolution map but still keep essential information [1].

\begin{figure}
\centering
\includegraphics[width=0.5\textwidth]{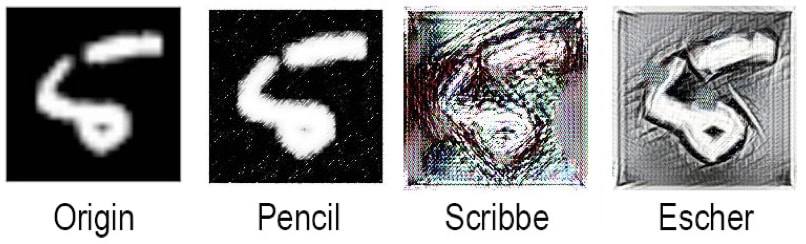}
\caption{\label{fig:illustration} Feature extraction with different filters for digit 5}
\end{figure}

Fig.~\ref{fig:illustration} depicts an illustration regarding digit 5 when using different feature extractions (filter effects), namely, Origin (where there is no filter applied), Pencil, Scribble and Escher \cite{lunapic2017}. As we can see, these filters remove background noise in the image and make the digit more or less easier to recognize.

\section{Results}
\label{sec:results}
In this section, we discuss results of three experiments including finding the best learning rates for MLP, comparison of MLPs and CNN and performance of detection on each digit.

\subsection{Experiment 1: Finding the Best Learning Rates}
Since Keras sets default learning rate at $0.5$, we modified this value to find the best learning rate for our dataset. We varied the values as $1$, $0.5$, $0.1$ and $0.01$ and performed iterations from $1$ to $10$ with $1$ increment. From Table \ref{tab:error}, we found that when the learning rate is $0.1$, the error rate is lowest ($0.05$ versus $5.1$, $1.03$, $2.67$ for learning rates of $1$, $0.5$ and $0.01$).

\begin{table}[htb!]
\centering
\caption{\label{tab:error}Error Rates for Different Learning Rates of MLP.}
\resizebox{\columnwidth}{!}{
\begin{tabular}{|r|l|l|l|l|}
\hline
\multicolumn{1}{|l|}{\multirow{2}{*}{\#Iteration}} & \multicolumn{4}{c|}{Error Rates (\%)}                                                                                \\ \cline{2-5} 
\multicolumn{1}{|l|}{} & \multicolumn{1}{r|}{lr=1} & \multicolumn{1}{r|}{lr=0.5} & \multicolumn{1}{r|}{lr=0.1} & \multicolumn{1}{r|}{lr=0.01} \\ \hline
1  & 10.74 & 3.46 & 3.31 & 8.38 \\ \hline
2  & 7.45  & 2.16 & 2.08 & 6.73 \\ \hline
3  & 7.66  & 2.21 & 1.31 & 5.71 \\ \hline
4  & 6.84  & 2.29 & 1.19 & 4.86 \\ \hline
5  & 8.80  & 1.69 & 0.64 & 4.44 \\ \hline
6  & 6.61  & 0.97 & 0.42 & 3.91 \\ \hline
7  & 4.82  & 0.80 & 0.37 & 3.56 \\ \hline
8  & 4.82  & 0.73 & 0.14 & 3.16 \\ \hline
9  & 6.36  & 0.67 & 0.08 & 2.88 \\ \hline
\cellcolor[rgb]{0.000,0.700,1.000}10 & \cellcolor[rgb]{0.000,0.700,1.000}5.10  & \cellcolor[rgb]{0.000,0.700,1.000}1.03 & \cellcolor[rgb]{0.000,0.700,1.000}0.05 & \cellcolor[rgb]{0.000,0.700,1.000}2.67 \\ \hline
\end{tabular}}
\end{table}
\subsection{Experiment 2: CNN vs MLPs comparison}
\begin{figure}
\centering
\includegraphics[width=0.5\textwidth]{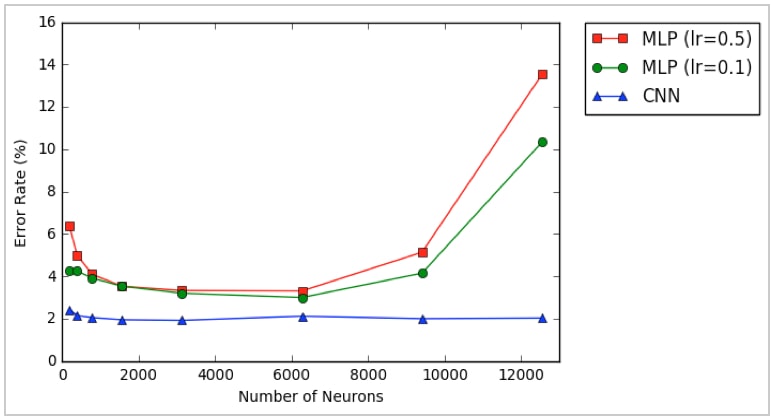}
\caption{\label{fig:comparison_cnn}Comparison of CNN and MLPs}
\end{figure}

In this experiment, we compared MLPs (with default learning rate and our best learning rate) and CNN. We varied the number of neurons in the input layer from 196 (¼ the size of the input vector) to 12544 (16 times the size of the input vector). Table \ref{tab:percentages} shows the results regarding the error rates. We also plot these results in Fig.~\ref{fig:comparison_cnn} for convenient comparison.

As it can be observed from Fig. \ref{fig:comparison_cnn}, CNN outperforms both MLPs with the default learning rate and the best learning rate. We also see that when MLPs start getting overfit (error rates begin to increase), CNN is still stable.

\begin{table}[htb]
\centering
\caption{\label{tab:percentages}Percentage of error rates for different numbers of input neurons for MLP and CNN.}
\resizebox{\columnwidth}{!}{
\begin{tabular}{|r|l|l|l|}
\hline
\multicolumn{1}{|l|}{\multirow{2}{*}{\#Number of input neurons}} & \multicolumn{3}{c|}{Error (\%)} \\ \cline{2-4} 
\multicolumn{1}{|l|}{} & MLP (lr=0.5) & MLP (lr=0) & CNN  \\ \hline
196   & 6.36  & 4.25  & 2.42 \\ \hline
392   & 5     & 4.24  & 2.14 \\ \hline
784   & 4.09  & 3.91  & 2.04 \\ \hline
1568  & 3.52  & 3.52  & 1.94 \\ \hline
3136  & 3.32  & 3.19  & 1.91 \\ \hline
6272  & 3.3   & 2.99  & 2.11 \\ \hline
9408  & 5.13  & 4.14  & 1.99 \\ \hline
12544 & 13.53 & 10.33 & 2.02 \\ \hline
\end{tabular}}
\end{table}

\subsection{Experiment 3: Performance of Recognition on individual Digits}
This experiment is designed to determine if our CNN can recognize a certain digit better than others. Table \ref{tab:precision} shows Precision, Recall and F1-Score of these digits. From this table, digit 0 can be recognized better than the others in terms of Precision. However, in terms of Recall and F1-Score, digit 1 performs better than other digits.

Notice that, these results are varied each time we change the weights of CNN (we initiate the CNN with different seeds). For example, if the seed is 7, the best F1-Score is on digit 1, but if seed is 17, the best F1-Score is on digit 4. So we may conclude that the performance of CNN on each digit depends on the initial weights of CNN.

\begin{table}[]
\centering
\caption{\label{tab:precision}Precision, Recall and F1-Score on Digits.}
\resizebox{\columnwidth}{!}{
\begin{tabular}{|r|l|l|l|}
\hline
\multicolumn{1}{|l|}{Digit} & Precision & Recall & F1-Score \\ \hline
0 & \cellcolor[rgb]{0.000,0.700,1.000}99.78 & 98.88 & 98.83 \\ \hline
1 & 98.86 & \cellcolor[rgb]{0.000,0.700,1.000}99.30 & \cellcolor[rgb]{0.000,0.700,1.000}99.08 \\ \hline
2 & 98.34 & 97.67 & 98.01 \\ \hline
3 & 97.54 & 98.22 & 97.88 \\ \hline
4 & 97.87 & 98.37 & 98.12 \\ \hline
5 & 98.20 & 97.87 & 98.03 \\ \hline
6 & 98.13 & 98.75 & 98.44 \\ \hline
7 & 98.53 & 97.67 & 98.09 \\ \hline
8 & 97.95 & 98.25 & 98.10 \\ \hline
9 & 98.00 & 97.22 & 97.61 \\ \hline
\end{tabular}}
\end{table}

\section{Online Implementation}
\label{sec:online}
Offline implementation has shown a promising approach in Deep Learning. However, it suffers from a drawback in terms of usability because the training process occurs inside local computers that makes accessing from outsiders difficult. Thus, the offline approach limits the number of users cooperating in Deep Learning projects. Recently, there is several attempts to bring Deep Learning for online production via web for example ConvNetJS and NeuroJS. ConvNetJS is written entirely in Javascript and supports Convolutional Neural Network and other Neural Networks. In the following, we present our experiments based on this ConvNetJS library.

\subsection{Cifar-10 Dataset}
\begin{figure}[htb!]
\centering
\includegraphics[width=0.5\textwidth]{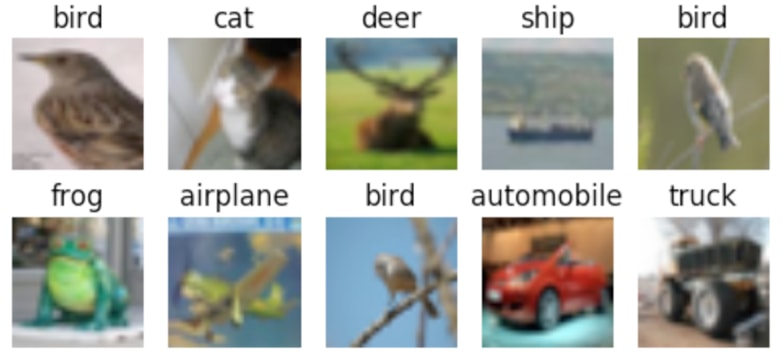}
\caption{\label{fig:random10}Visualization of random images Cifar-10}
\end{figure}

We performed these experiments based on Cifar-10 which offers $60000$ colour images with the size of $32\times 32$ pixels. There are 10 categories including airplane, automobile, bird, cat, deer, dog, frog, horse, ship and truck with $6000$ images in each category. All images are bundled into a training of $50000$ images and testing set of $10000$ images. These training images are grouped into 5 batches of $1000$ images beside of 1 batch for testing images. The testing batch contains exactly $1000$ images of each class where as a training batch may contains more or less than $1000$ images of each class \cite{krizhevsky2009learning}. Fig.~\ref{fig:random10} shows 10 random images because of limitation in the number of pages.

\subsection{Experiment 4}
This experiment is designed to compare performances of a Convolutional Neural Network with regard to several optimizers. The network mainly consists of two convolutional layers and a fully connected network. The first convolutional layer involves 16 filters have the size of $5\times 5$ and followed by a pool with the size of $2\times 2$. The second convolutional layer comprises of 20 filters and a pool with the same settings as the first layer.

\begin{figure}[tbh!]
\centering
\includegraphics[keepaspectratio, width=8cm]{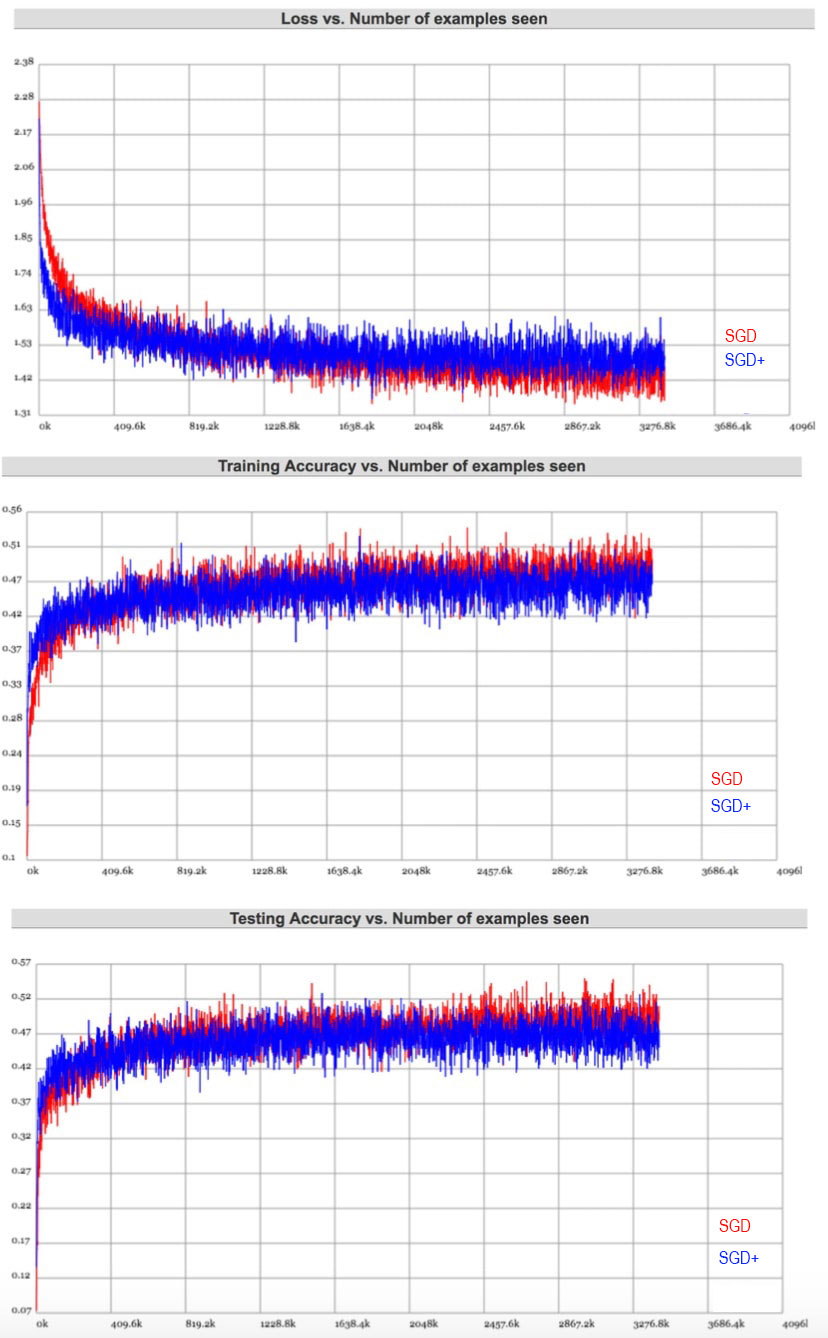}
\caption{\label{fig:comparison_sgd}Comparisons of SGDs in Cifar-10}
\end{figure}

ReLU activation is used in both layers. The output layer is set to classify the 10 different categories. We perform comparisons of two SGD optimizers with the default learning rate and two variations of momentum 0.0 and 0.9. We named these optimizers as SGD and SGD+, respectively. Results of the loss function, training and testing accuracies on number of samples are shown in Fig. \ref{fig:comparison_sgd}. As it can be depicted from the figure, the performances of default SGD are sometimes greater than the performances of SGD+, but at other times are less than those of SGD+. When the number of samples is small (approximately $400000$ samples), SGD+ performs better across Loss, Training accuracy and Testing accuracy domains. However, SGD performs generally better with more samples. In addition, accuracies in training and testing are increasing steadily despite slow rates (accuracy achieves roundly $0.5$ after $2000$k samples).

\subsection{Experiment 5}
\begin{figure}[htb!]
\centering
\includegraphics[width=0.5\textwidth]{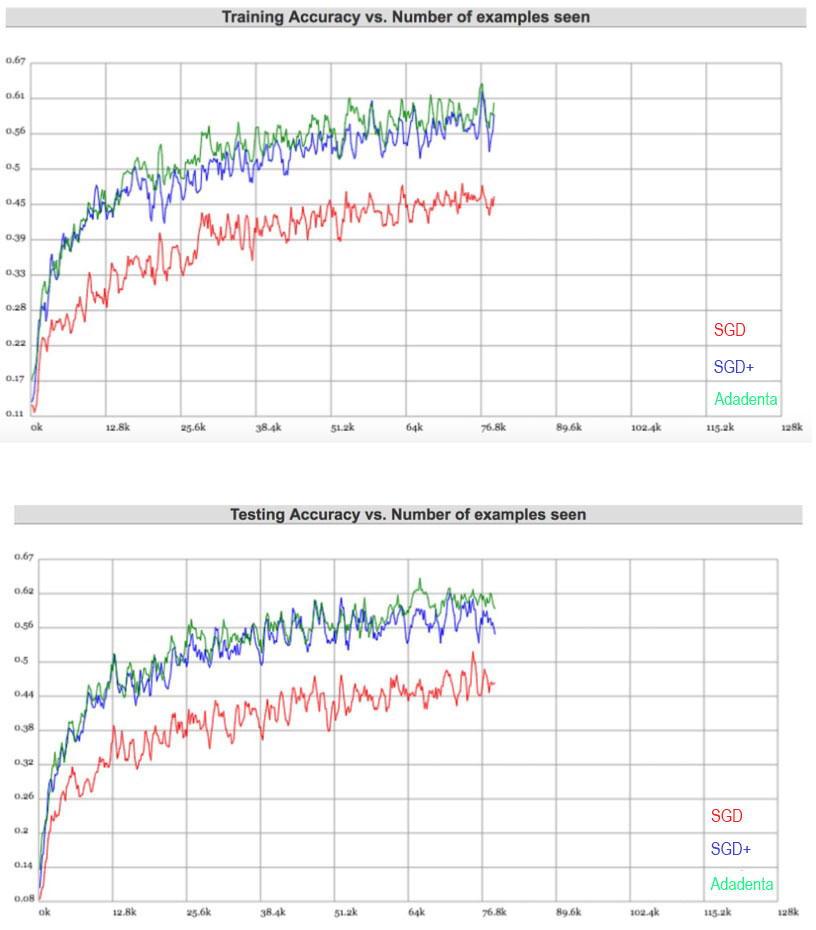}
\caption{\label{fig:comparison_optimizers}Comparison of different optimizers in Cifar-10}
\end{figure}

In this experiment, we tried to improve the accuracy of the Convolutional Neural Network by varying different architectures. We added a dropout in second convolutional layer and also an Adadelta optimizer. Results of this experiment are plotted in the Fig. \ref{fig:comparison_optimizers}. We can observe that performances are improved significantly. In training set, SGD+ can accomplish the accuracy of $0.6$ after about $70$k samples (versus $2000$k in previous setting). In the same time, Adadelta performs a slightly better than SGD+. We also see similar trends for testing set.

\section{Conclusion and Future Work}
\label{sec:conclusion}
In this research, we present our findings for offline and online deep learning in image recognition. In offline recognition, we setup a deep learning development environment built around TensorFlow and Keras. We also performed comparisons of different optimizers. Results showed that CNN achieves more accuracy and is more stable than typical fully-connected networks. In addition, performances are varied across digits. In online image recognition, we setup a web-based approach surrounding a Javascript library for deep learning. Several optimizers were tested and Adadelta slightly outperforms the best SGD in our setting.

We argue that though using Convolutional Neural Network does not require expert's knowledge, handcraft filters may result in a better performance since classifying certain objects may indeed require a more concrete understanding of a typical field. Besides of handcraft filters, we also plan to further improve the performance of the optimizers.

\bibliographystyle{ieeetr}
\bibliography{online}

\end{document}